\pdfoutput=1

\documentclass[11pt]{article}

\usepackage{acl}

\usepackage{times}
\usepackage{latexsym}
\usepackage{amsmath}
\usepackage{amssymb}
\usepackage{textcomp}
\usepackage{array}
\usepackage{tikz}
\usepackage{multirow}
\usepackage{booktabs}
\usepackage{hyperref}
\usepackage{natbib}

\usepackage[T1]{fontenc}

\usepackage[utf8]{inputenc}

\usepackage{microtype}

\newcolumntype{P}[1]{>{\centering\arraybackslash}m{#1}}

\title{Detecting Euphemisms with Literal Descriptions and Visual Imagery}

\author{
Ilker Kesen\textsuperscript{1,2} \quad
Aykut Erdem\textsuperscript{1,2}\quad
Erkut Erdem\textsuperscript{1,3}\quad
Iacer Calixto\textsuperscript{4,5}
\\
\textsuperscript{1} Ko\c{c} University, KUIS AI Center ~\textsuperscript{2} Ko\c{c} University, Computer Engineering Department \\
\textsuperscript{3} Hacettepe University, Computer Engineering Department~ \\
\textsuperscript{4} Amsterdam UMC, University of Amsterdam, Department of Medical Informatics \\
\textsuperscript{5} Amsterdam Public Health, Methodology \& Mental Health, Amsterdam, The Netherlands
}

\begin{document}

\maketitle
\begin{abstract}
This paper describes our two-stage system\footnote{Code is available at \href{http://github.com/ilkerkesen/euphemism}{github.com/ilkerkesen/euphemism}} for the Euphemism Detection shared task hosted by the 3rd Workshop on Figurative Language Processing in conjunction with EMNLP 2022. Euphemisms tone down expressions about sensitive or unpleasant issues like addiction and death. The ambiguous nature of euphemistic words or expressions makes it challenging to detect their actual meaning within a context.
In the first stage, we seek to mitigate this ambiguity by incorporating literal descriptions into input text prompts to our baseline model. It turns out that this kind of direct supervision yields remarkable performance improvement. In the second stage, we integrate visual supervision into our system using \textit{visual imageries}, two sets of images generated by a text-to-image model by taking terms and descriptions as input. Our experiments demonstrate that visual supervision also gives a statistically significant performance boost. Our system achieved the second place with an F1 score of 87.2\%, only about 0.9\% worse than the best submission.
\end{abstract}

\section{Introduction}
Recent advances in
large %
pretrained language models allowed the computational linguistics community to tackle more
knowledge-intensive %
tasks which require commonsense reasoning \citep{talmor-etal-2019-commonsenseqa,bisk2020piqa,lin-etal-2021-differentiable}, %
and figurative language understanding \citep{pedinotti-etal-2021-howling,liu-etal-2022-testing}. In this work, we focus on a figurative language understanding task called \textit{euphemism detection}. Euphemisms attempt to smooth harsh, impolite, or blunt expressions about taboo or sensitive topics like death and unemployment \citep{holder2008dictionary}. For instance, when we speak of older people we often refer to \textit{senior citizens} instead of a direct expression that can be seen as offensive.

Identifying euphemisms is
challenging due to their natural ambiguity, i.e., the meaning of the term shifts depending on the context: `\textit{Over the hill}' could either mean someone or something is \textit{physically} over some hill (\emph{literal}), or someone or something is \textit{old}, past one's prime (\emph{figurative}) \citep{lee_searching_2022}. One cannot distinguish these
two different senses without sufficient context. Thus, these terms are referred as \textit{potentially euphemistic terms} (PETs) \citep{gavidia_cats_2022}. Here, we propose a two-stage method for the Euphemism Detection shared task hosted by the 3rd Workshop on Figurative Language Processing at EMNLP 2022. 

In the first stage, we manually collect literal descriptions for each PET. We then incorporate these descriptions into input text prompts to help the model distinguish figurative from literal usage. We demonstrate that this kind of extraneous linguistic supervision improves a strong baseline by a large margin. In the second stage, we attempt to answer the question, \textit{``Is visual supervision also useful to
infer the %
meaning behind a PET?''} To answer this question, we use a text-to-image model which takes terms and descriptions as input, and we generate two sets of images, which we denote as \textit{visual imageries}. Our experiments show that using visual imagery provides the best results. A paired t-test points out that the improvement is statistically significant. Our qualitative analysis also suggests visual imageries are beneficial for analyzing PETs.

The rest of this paper is organized as follows. Section~\ref{sec:approach} describes our proposed solution. In Section~\ref{sec:data-implementation}, we share the details of our evaluation~setup and design choices. Section~\ref{sec:results} reports our experimental results. In Section~\ref{sec:related-work}, we briefly review the relevant literature. Section~\ref{sec:conclusion} outlines our conclusions and discuss the limitations of our approach.

\section{Approach}
\label{sec:approach}
In this section, we first formulate the euphemism detection task by describing a simple baseline model, and then explain how we extend it with the literal term descriptions and visual imagery.

\subsection{Vanilla Baseline}
\label{sec:vanilla-baseline}
Given a textual context $C$ with a potentially euphemistic term (PET) $T$, the aim of euphemism detection is to decide whether the candidate term $T$ is euphemistic ($y=1$) or not ($y=0$). Here, we only pick a sentence $S = [w_1, w_2, ..., w_n]$ which contains
a %
candidate term $T$, and ignore the rest of the context $C$ at first. We use a pretrained language model LM as our initial baseline as below.
\vspace{-0.25cm}

\begin{align*}
\begin{array}{lll}
e_i &= \mbox{EMBED}(w_i), &
\multirow{2}{*}{$\hat{y} =
    \begin{cases} 
        1 & \hat{p}\geq 0.5, \\
        0 & \text{otherwise}.
    \end{cases}$}\\
\hat{p} &= \mbox{LM}(e_1, e_2, ..., e_n),\\
\end{array}
\end{align*}
$e_i$ denotes the word embedding of the i\textsuperscript{th} token
$w_i$, $\hat{p}$ is the probability that the candidate term $T$ is euphemistic, and $\hat{y}$ is the predicted label. $\mbox{EMBED}$ is the embedding layer and $\mbox{LM}$ denotes the language model that produces the probability $\hat{p}$.

\subsection{Literal Descriptions}
\label{sec:literal-descriptions}
We extend the baseline model by supplying extra supervision with literal descriptions $D$ for each candidate term $T$ (which we collect manually). To make use of the literal descriptions, we create a textual prompt $X = [x_1, x_2, ..., x_n]$ for each sentence $S$, term $T$ and description $D$
as below.
\vspace{0.05cm}
\begin{equation*}
    X = [\text{Term:}~T,~~\text{Description:}~D,~~ \text{Sentence:}~S].
\end{equation*}
Then, we change the formulation,
\vspace{-0.05cm}
\begin{align*}
    e_i &= \mbox{EMBED}(x_i) \\
    \hat{p} &= \mbox{LM}(e_1, e_2, ..., e_n),
\end{align*}
\noindent
where $e_i$ is the embedding for the i\textsuperscript{th} token of the input prompt $X$. 

\subsection{Visual Imagery}
\label{sec:visual-imagery}
We subsequently move beyond the text-only baselines by integrating visual modality into the Literal Descriptions baseline in the form of \textit{visual imagery}. To accomplish this, we generate two sets of images $I_T = [I_T^{(1)}, I_T^{(2)}, ..., I_T^{(k)}]$ and $I_D = [I_D^{(1)}, I_D^{(2)}, ..., I_D^{(k)}]$, for each term and description pair, respectively. We denote these set of images as visual imageries. To obtain the visual imageries, we feed a text-to-image model $\mbox{T2I}$ with terms and descriptions as input language,
\begin{align*}
    I_T^{(k)} &\sim \mbox{T2I}(T), & I_D^{(k)} &\sim \mbox{T2I}(D).
\end{align*}

Next, we use a pretrained visual encoder ($\mbox{VE}$) to embed visual imageries.
\begin{align*}
    v_T &= \frac{1}{K} \sum_{k=1}^{K} \mbox{VE}(I_T^{(k)}), &
    v_D &= \frac{1}{K} \sum_{k=1}^{K} \mbox{VE}(I_D^{(k)})
\end{align*}
where $v_T$ denotes the visual imagery embedding of the candidate term $T$ and $v_D$ denotes the visual imagery embedding of the corresponding literal description $D$.
$K$ is the number of images per term $T$ and description $D$.
Thus, we reformulate the literal descriptions baseline as follows,
\begin{align*}
    e_i &= \mbox{EMBED}(x_i)\\
    \hat{p} &= \mbox{LM}(f_p(v_T), f_p(v_D), e_1, e_2, ..., e_n)
\end{align*}
We make sure visual imagery embeddings are compatible with the word embeddings and language model LM by applying a linear projection layer $f_p$.
We train each baseline using the negative log-likelihood objective.

\section{Data and Implementation}
\label{sec:data-implementation}

\paragraph{Data.} The euphemism detection dataset consists of two separate splits for training and testing purposes
with 1573 and 394 examples, respectively.
The test split is unlabeled. The whole data includes 131 different PETs. Since there is no data supplied for validation, we reserve 20\% of the training data for this purpose.
We only select the sentences with PETs and remove repetitive patterns of punctuation \textit{"@ @ @ ..."} to decrease computational requirements by shortening the input language. We manually collect literal descriptions within 6 hours, and try to avoid impolite expressions like insults or slang phrases.

\paragraph{Implementation.}
We use 
DeBERTa-v3 base and large as our language model \citep{he2021debertav3,he2021deberta}. We generate the visual imageries $I_T$ and $I_D$ by using an open-source DALL-E implementation~\citep{pmlr-v139-ramesh21a,Dayma_DALL-E_Mini_2021}.\footnote{\href{https://github.com/kuprel/min-dalle}{github.com/kuprel/min-dalle}} The number of images per visual imagery $K$ is set to 9. We extract visual imagery embeddings $v_T$ and $v_D$ using CLIP's ViT-L/14 as our visual encoder \citep{radford2021learning}. $f_p$~is a single linear layer, and we randomly initialize its weights. We use Adam optimizer with weight decay \citep{DBLP:journals/corr/KingmaB14,loshchilov2018decoupled}. The learning rate is set to $5e^{-6}$ and $3e^{-6}$ for the experiments with DeBERTa-v3-base and DeBERTa-v3-large, respectively. We train our models for a maximum of 50 epochs using Tesla V100s and mixed precision.
A typical experiment takes less than one hour with a batch size of 16. Due to
the small dataset size,
we perform multiple experiments
and reserve
a different portion of the labeled data for
validation in each experiment.
We report mean and standard deviation
over %
all experiments,
and use ensembling to evaluate our system on the test set.

\section{Experimental Analysis}
\label{sec:results}
\begin{table}[!t]
\renewcommand{\arraystretch}{1.1}
\centering
\resizebox{\linewidth}{!}{
\begin{tabular}{llcc}
\toprule
Model & LM & \textit{validation} & \textit{test} \\ \midrule 
Vanilla Baseline & Base & 79.84 \textpm 2.23 & - \\ 
+ Desc. & Base & 86.39 \textpm 1.05 & 83.58 \\
+ Desc. & Large & \underline{88.89} \textpm 1.35 & \underline{85.74} \\
+ Desc. + Imag. & Large & \textbf{90.11} \textpm 1.59 & \textbf{87.16} \\
\bottomrule
\end{tabular}
}
\caption{Quantitative results on the labeled data using F1 as evaluation metric. The last two columns respectively show the average score over different \textit{validation} splits, and the ensemble performance achieved on the \textit{test} split.}
\label{table:main-results}
\end{table}

\subsection{Quantitative Results} 
Table \ref{table:main-results} presents the quantitative results of our experiments as ablation studies. We perform several experiments in a curriculum, where each following experiment activates a different feature (e.g. literal descriptions). We first implement a vanilla baseline using DeBERTa-v3-base, which lacks descriptions and imagery.

\noindent \textbf{Using Literal Descriptions.} In our first ablative analysis, we incorporate the literal term descriptions into the vanilla baseline described in Section \ref{sec:literal-descriptions}. Integrating this
supervision results in
substantial
performance improvement,
i.e.
$\approx\!6.5$ points using F1 as evaluation metric.

\noindent \textbf{Larger Language Model.} We implement the literal descriptions model using a larger language model which is the large architecture of the DeBERTa-v3 model. Using a bigger $\mbox{LM}$ gives 2 points performance improvement.

\noindent \textbf{Visual Imagery.} We
now report on
the visual imagery model explained in Section \ref{sec:visual-imagery}. This model additionally uses two different visual embedding vectors, denoted as visual imageries, which are generated by a text-to-image model using terms and descriptions. By using this
extra visual supervision, we obtain 1.22 and 1.42 F1 score increments in validation and testing phases. A paired t-test is applied to determine the significance of the results: We obtained a p-value of $0.032$, which points out that this improvement is statistically significant ($p\!<\!0.05$).

\subsection{Qualitative Analysis}
\begin{figure}[!t]
\renewcommand{\arraystretch}{1.0}
\centering
\resizebox{\linewidth}{!}{
\begin{tabular}{>{\raggedright}P{2.0cm}>{\raggedright}P{2.5cm}P{3.5cm}P{3.5cm}}
\toprule
\Large{\textbf{Term}} & \Large{\textbf{Description}} & \Large{$\boldsymbol{I_T}$} & \Large{$\boldsymbol{I_D}$}\\
\midrule
\Large{late} & \Large{old person, elderly} &
\includegraphics[scale=0.36]{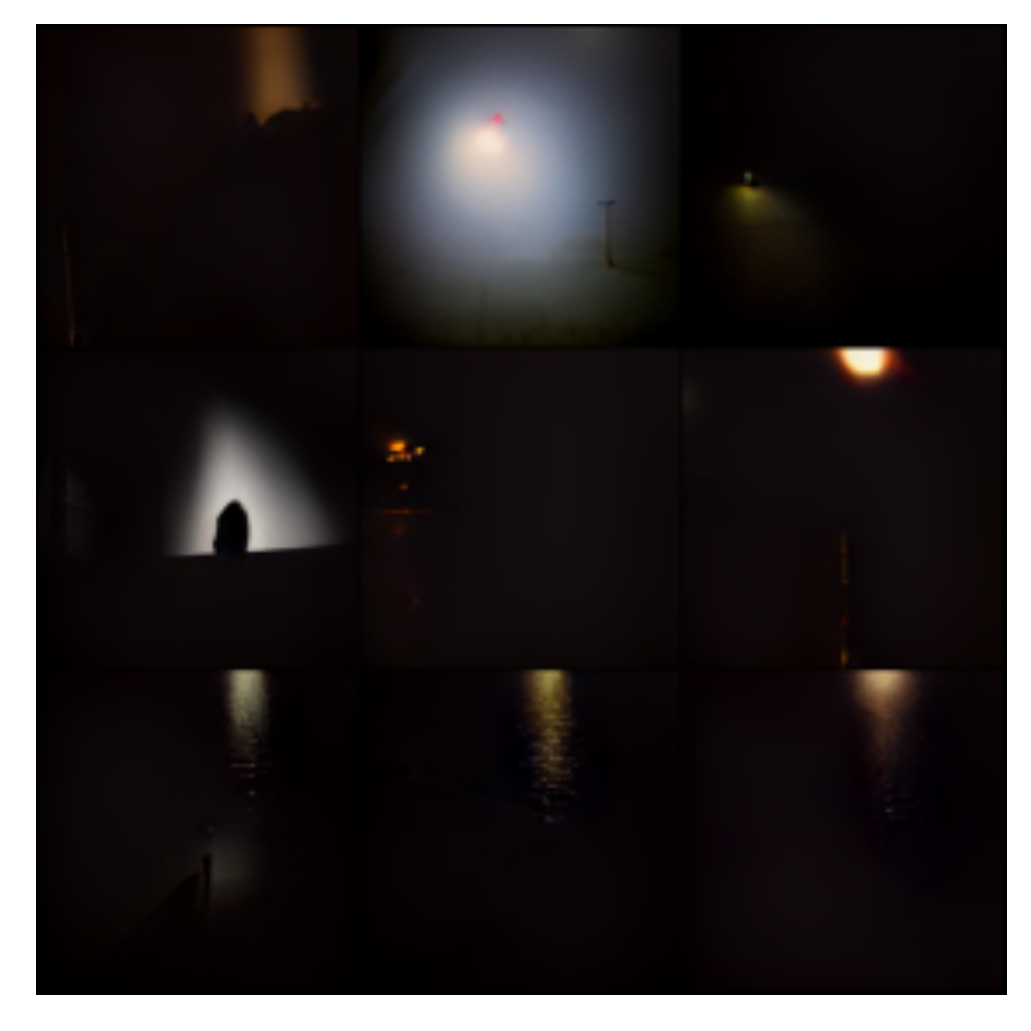} & \includegraphics[scale=0.36]{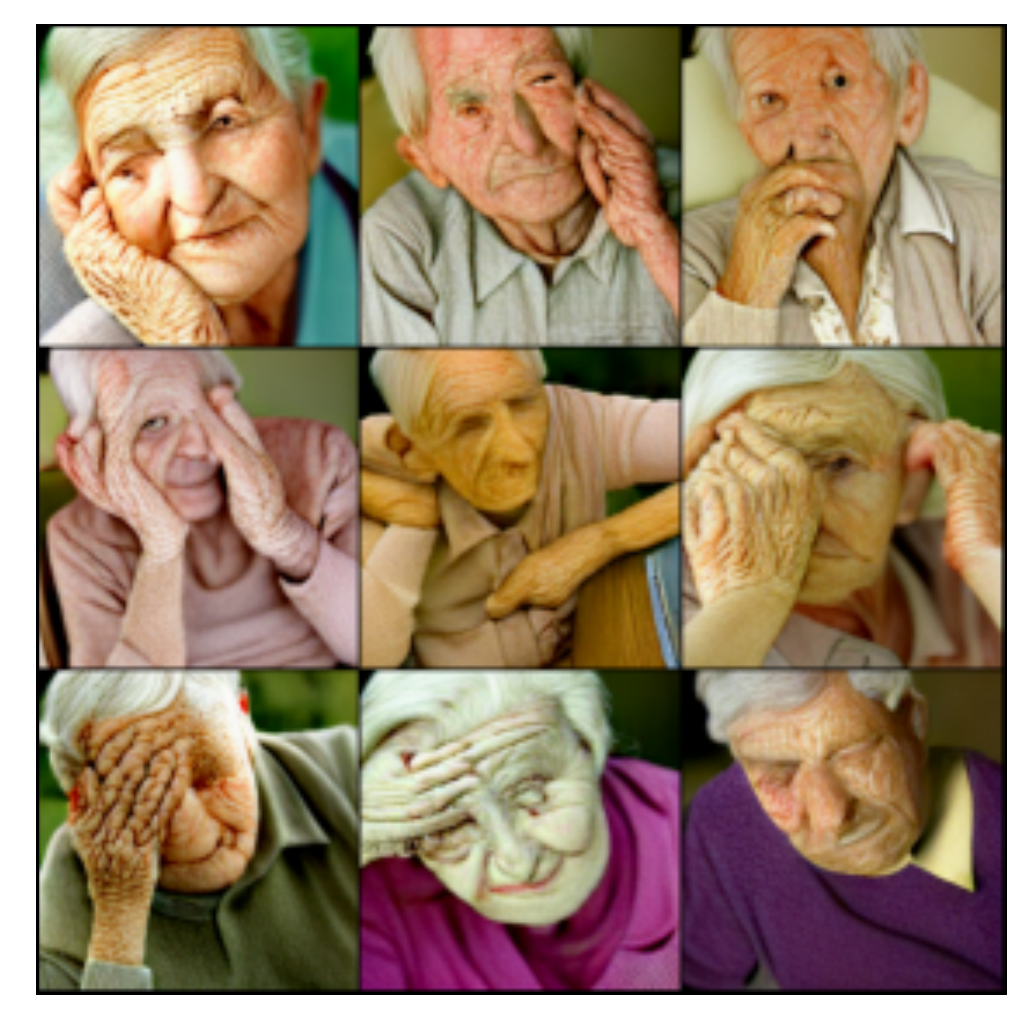}\\

\Large{pass} on & \Large{death, dying} &
\includegraphics[scale=0.36]{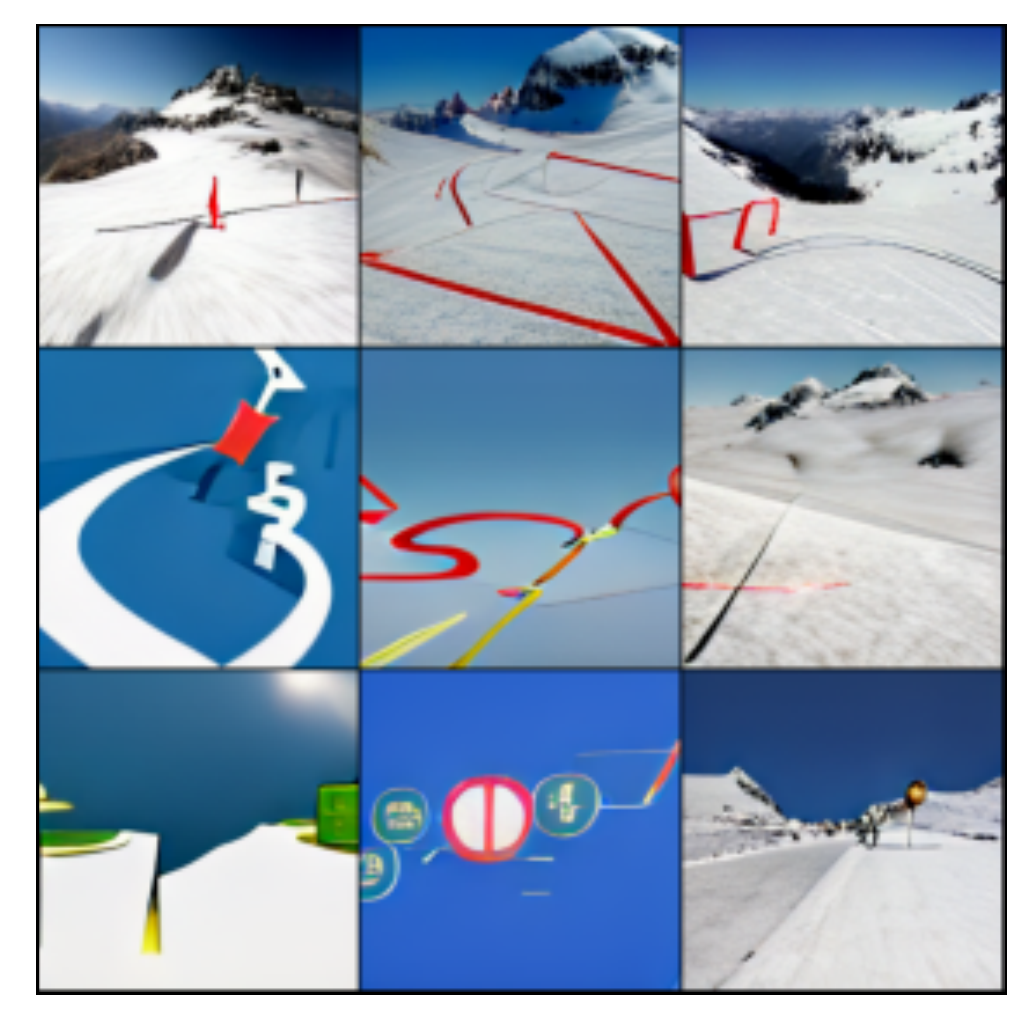} & \includegraphics[scale=0.36]{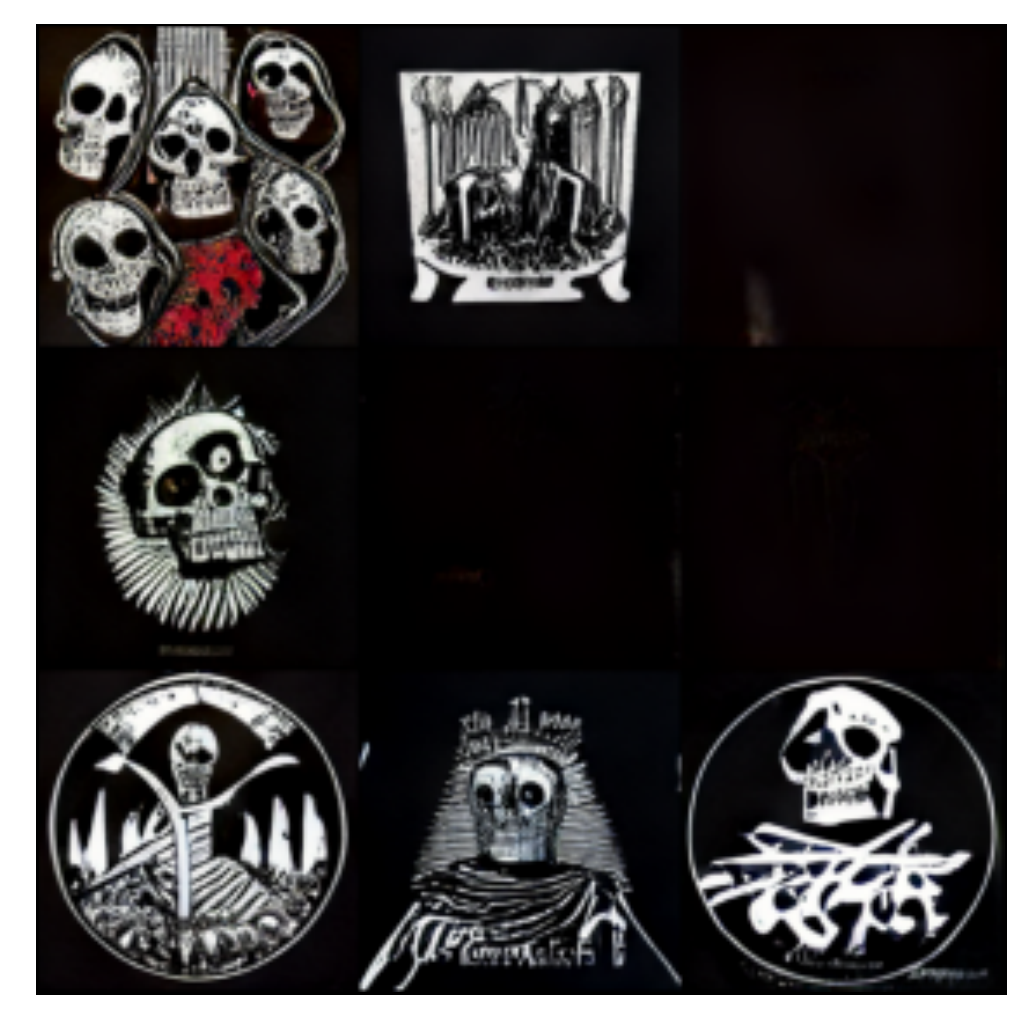}\\
 
\Large{lose one's lunch} & \Large{vomit, vomiting, throwing up} & \includegraphics[scale=0.36]{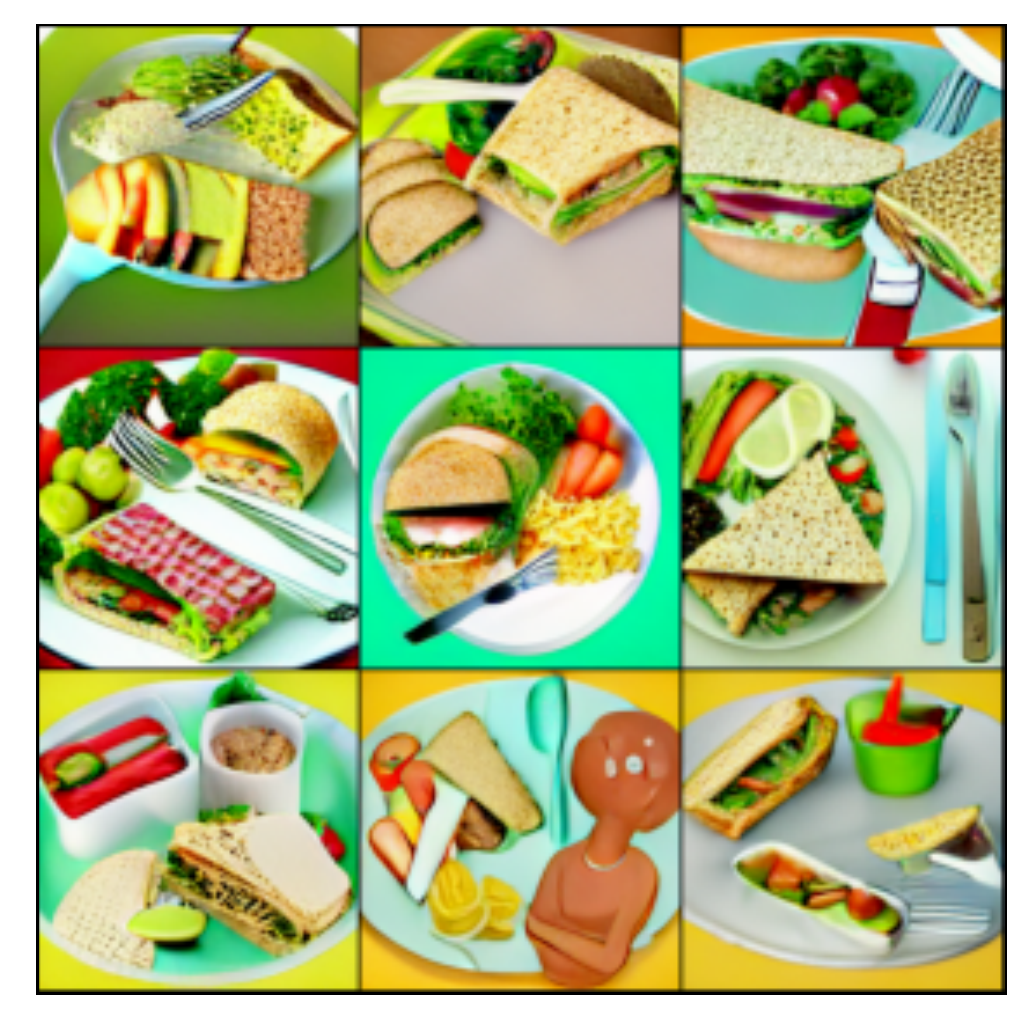}&\includegraphics[scale=0.36]{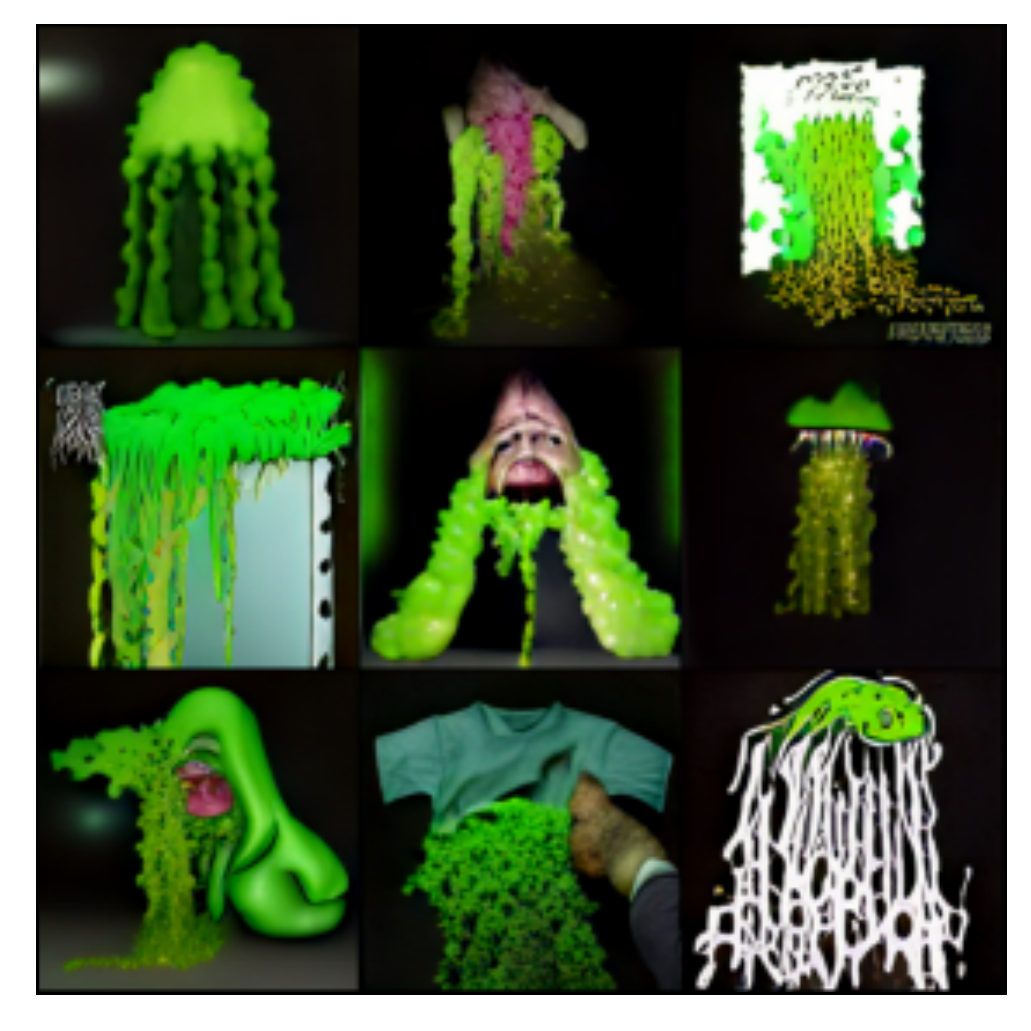}\\ 

\Large{pro-life} & {\Large{a person opposes abortion}} & \includegraphics[scale=0.36]{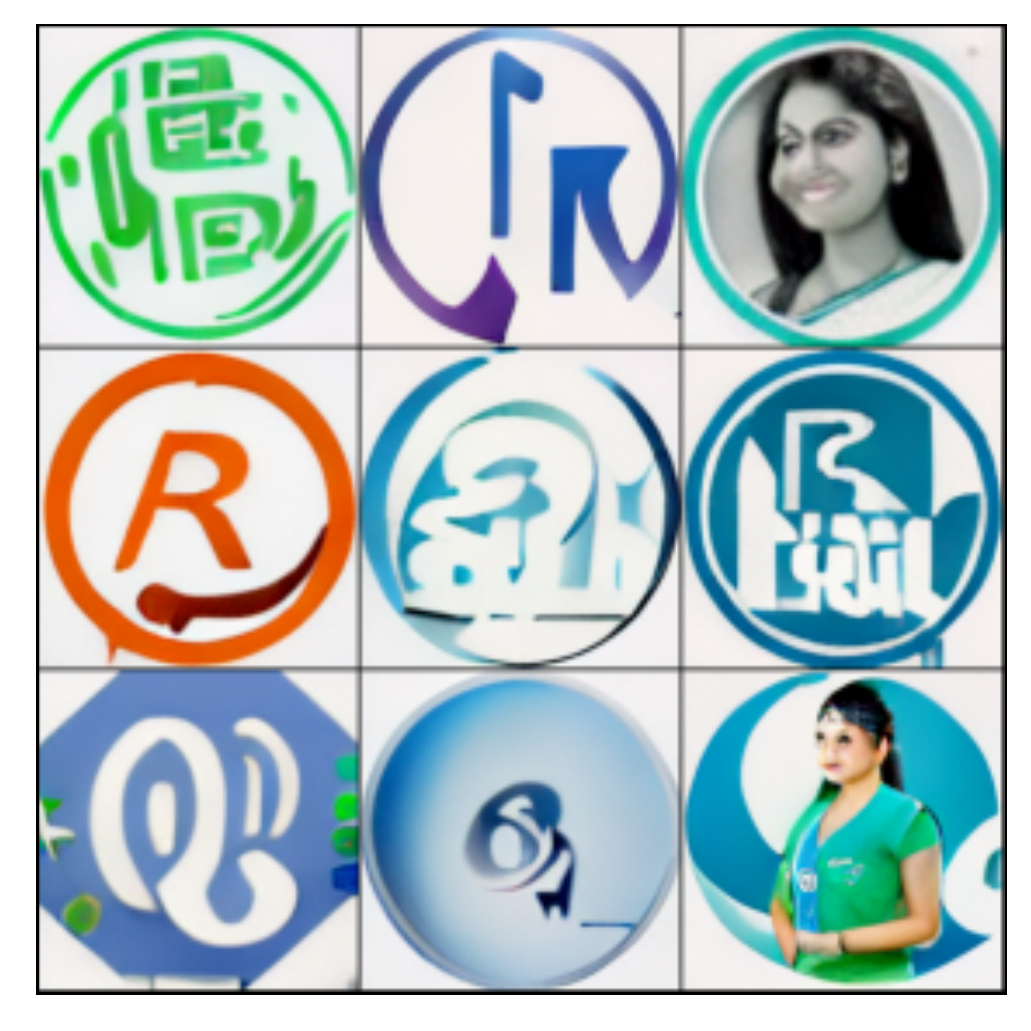}&\includegraphics[scale=0.36]{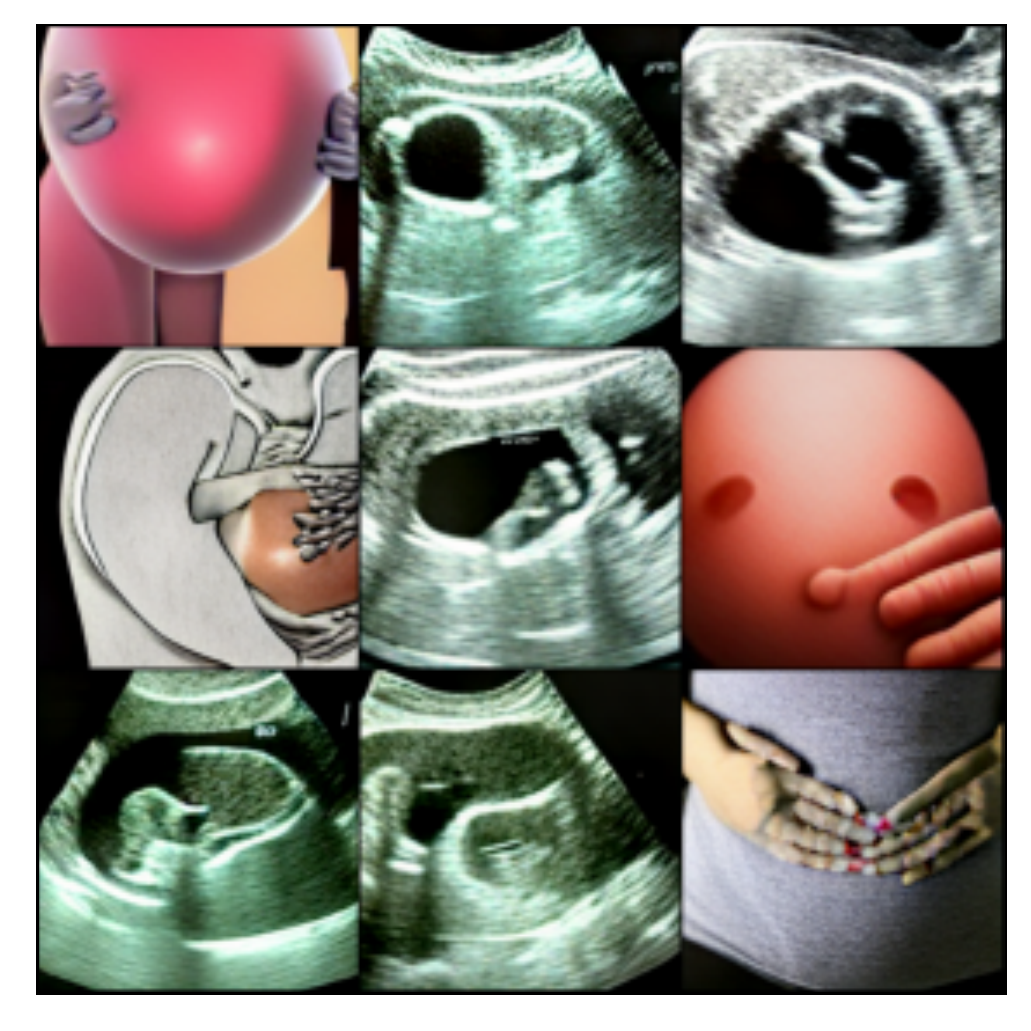}\\

\Large{able-body} & \Large{not disabled} & \includegraphics[scale=0.36]{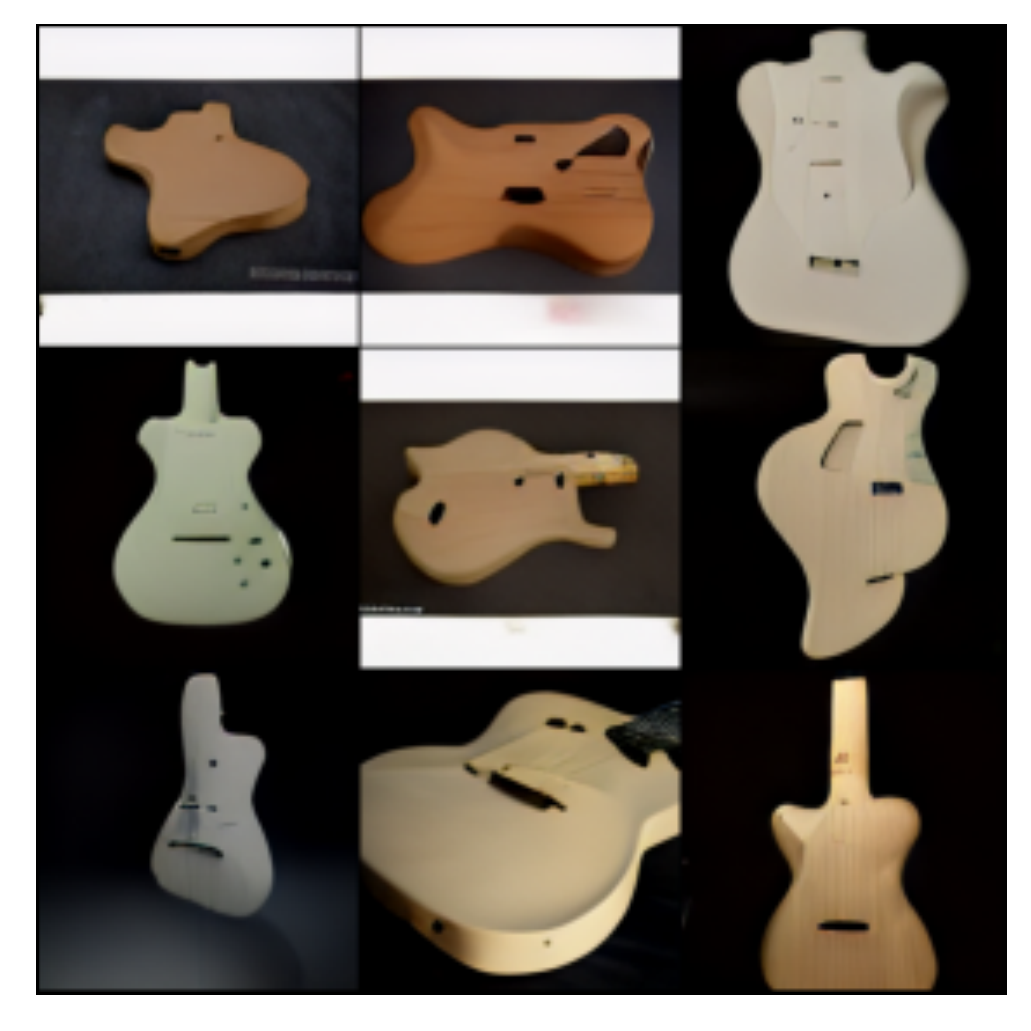}&\includegraphics[scale=0.36]{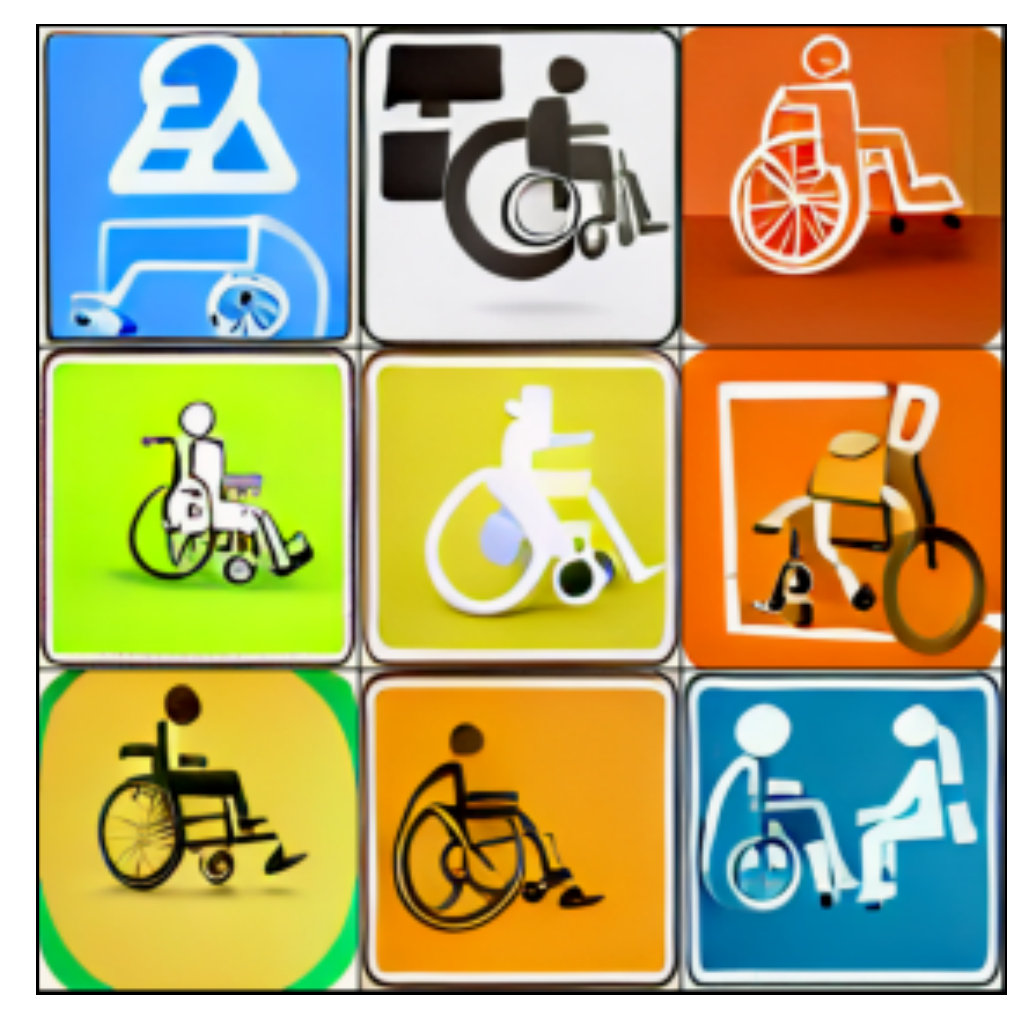}\\

\Large{lavatory} & \Large{restroom, toilet} & \includegraphics[scale=0.36]{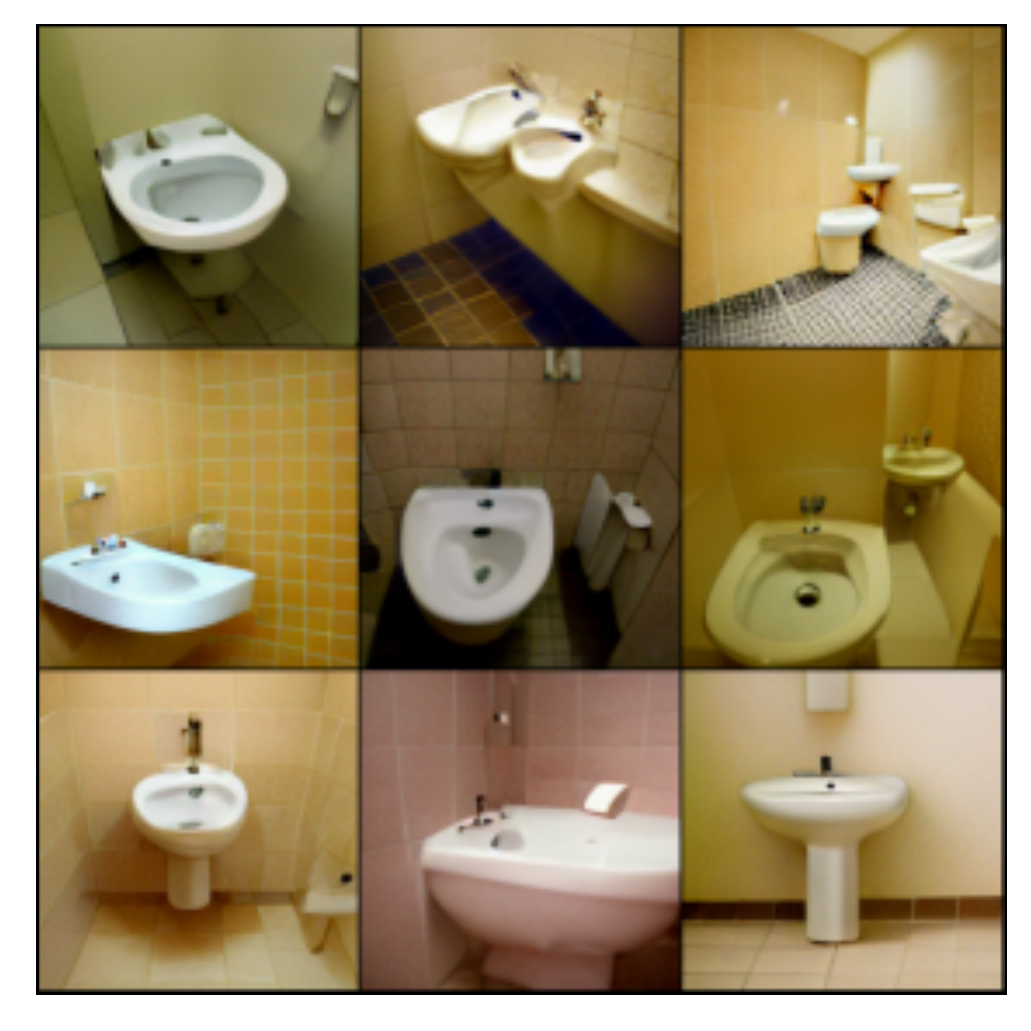}&\includegraphics[scale=0.36]{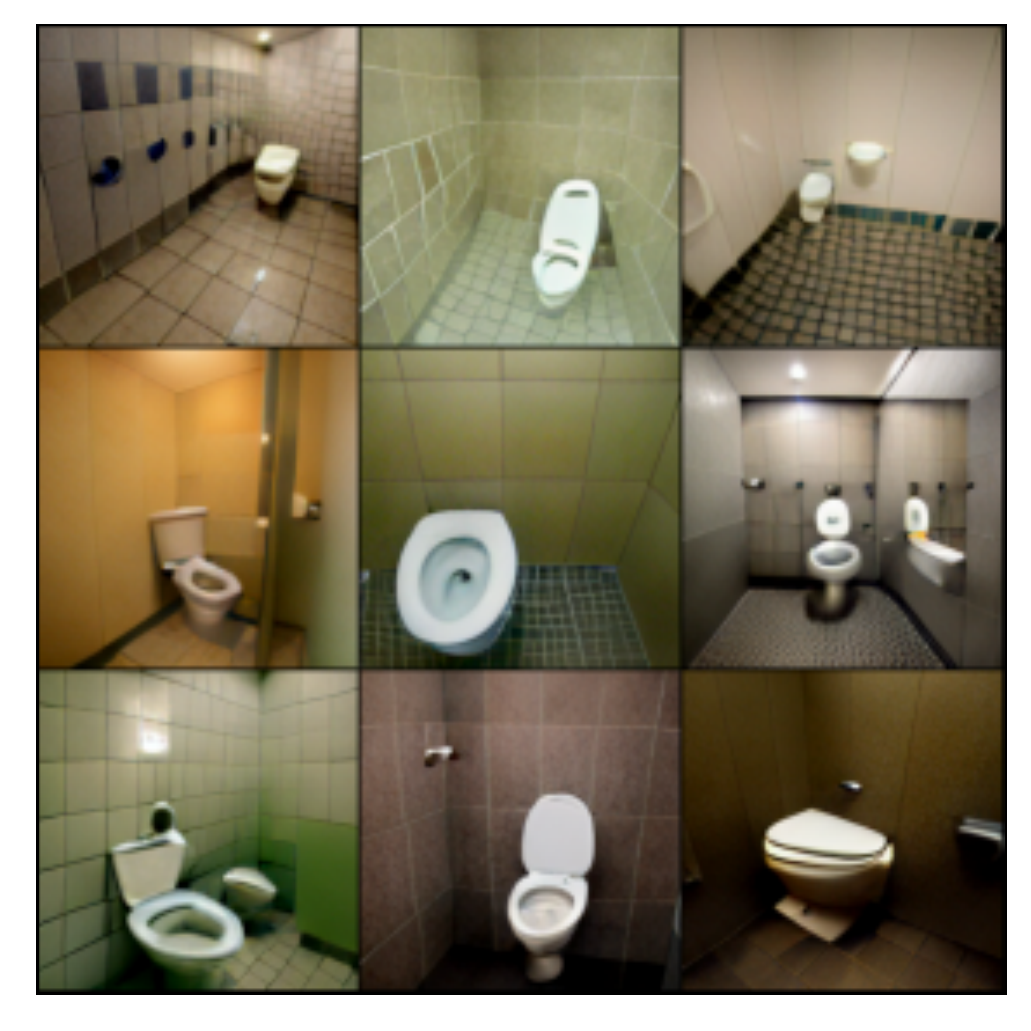}\\

\Large{senior citizen} & \Large{old person, elderly} & \includegraphics[scale=0.36]{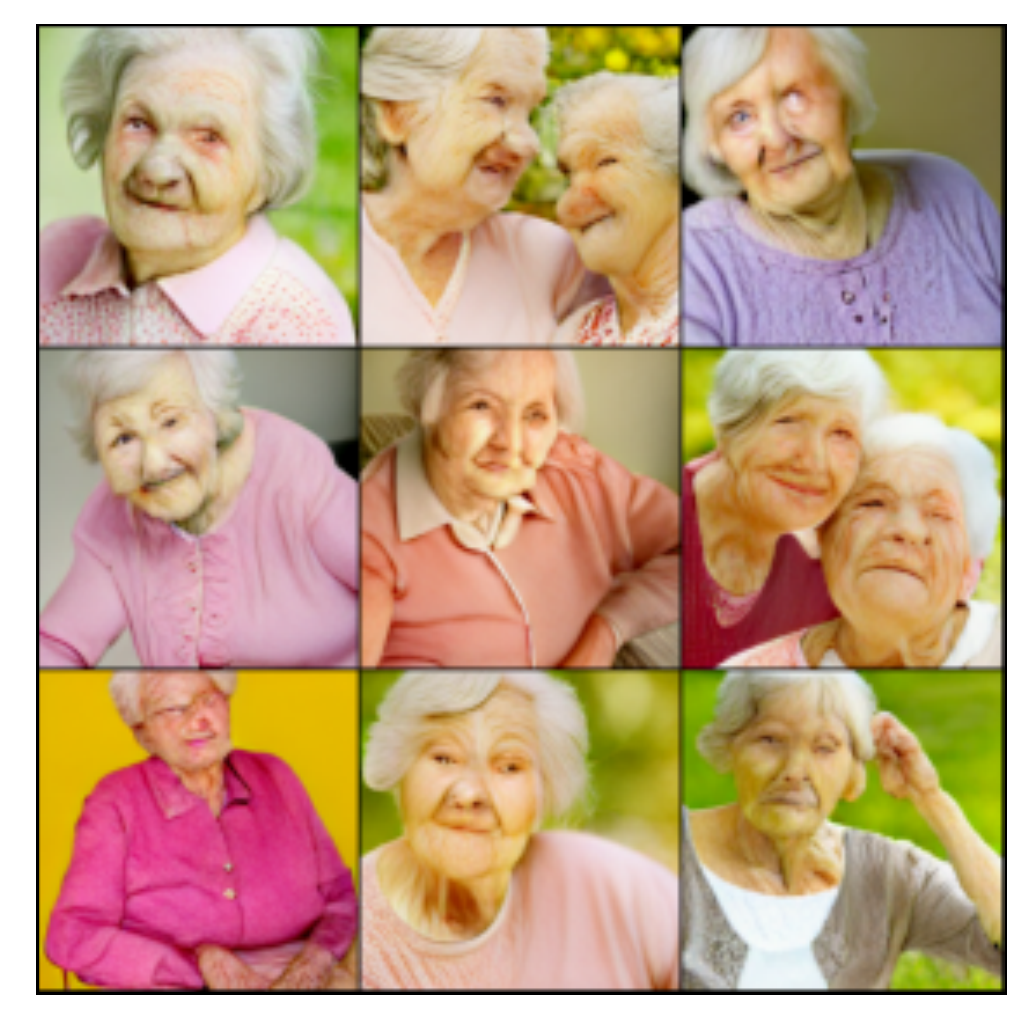}&\includegraphics[scale=0.36]{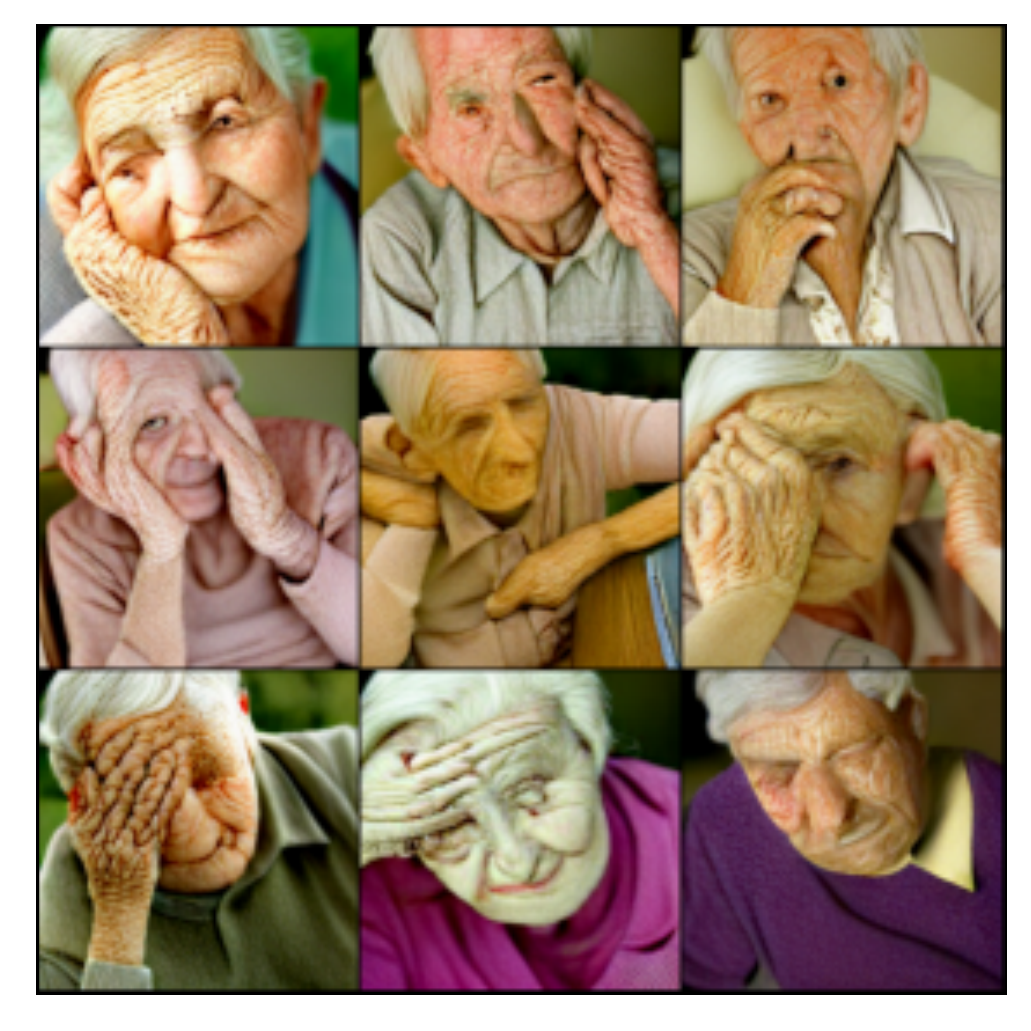}\\
\bottomrule
\end{tabular}
}
\caption{Examples of collected literal descriptions for euphemistic terms and their visual imageries.}
\label{table:visual-imageries}
\end{figure}

Figure \ref{table:visual-imageries} wraps up our qualitative analysis, where we share the collected descriptions and the generated visual imageries for some euphemistic terms. The first two examples show that if a term has a dominant literal meaning, the text-to-image $\mbox{V2I}$ model produces images conveying the literal meaning instead of the figurative one. $\mbox{V2I}$ can also produce visuals based upon individual word meanings as a consequence of being completely unconscious to the figurative meaning. This can be seen on the third example, where the model generates \textit{lunch} images instead of vomiting for phrase `\textit{lose one's lunch}'. Moreover, $\mbox{V2I}$ can generate unrelated images for some terms as one can see on the \textit{pro-life} and \textit{able-body} examples. On the other hand, the text-to-image model $\mbox{V2I}$ is well aware of some euphemism candidates as in the case with the last two examples. This phenomenon arises when the term has just one single meaning which is euphemistic.

In summary, a text-to-image model can be a complementary tool for analyzing figurative language: one can observe how models process these expressions. By looking at the produced images, we can recognize the terms with dominant literal meanings (e.g. \textit{late}) or single euphemistic meaning (e.g. \textit{lavatory}).

\section{Related Work}
\label{sec:related-work}
\noindent \textbf{Euphemisms.} Recently, euphemisms have attracted the attention of the natural language processing community.  \citet{zhu2021selfsupervised} and \citet{zhu-bhat-2021-euphemistic-phrase} extract euphemistic phrases by using masked language modeling. A few work practices sentiment-oriented methods to recognize candidate euphemism phrases \citep{felt-riloff-2020-recognizing,gavidia_cats_2022,lee_searching_2022}. Most notably, \citet{gavidia_cats_2022} replace PETs with their literal meanings and observe how the sentiment scores change. They demonstrate that using literal meanings produces higher scores for offensive speech and negative sentiment. Similarly, we also put literal meanings to use, but differently, by creating a textual input prompt. In this work, we also use the euphemism dataset they created.

\noindent \textbf{Knowledge-augmented Language Understanding.} External knowledge\footnote{Please check \citet{zhu-etal-2022-knowledge} for a comprehensive review of the related literature.} can be either unstructured (i.e. text) or structured (i.e. graph). To benefit from unstructured knowledge, a text retriever collects related entries from an external corpus \citep{karpukhin-etal-2020-dense,guu2020retrieval}. Conversely, structured knowledge integration may happen in two ways: explicit methods prefer to use knowledge in their input \citep{weijie2019kbert,zhang2019ernie}, and implicit methods try to learn knowledge in their objective \citep{xiong2019pretrained,shen-etal-2020-exploiting}. Some exceptions \citep{Yu_Zhu_Yang_Zeng_2022,lv2020commonsense} combines both: they learn to predict graph embeddings and use these embeddings as input in their model concurrently. Similar to us, \citet{yu-etal-2022-dict,xu-etal-2021-fusing,chakrabarty_figurative_2021} also insert descriptions into their textual inputs.

\noindent \textbf{Visually-aided Language Understanding.} Several methods have been proposed to aid language learning with external visual knowledge. Most of these methods experiment on machine translation (MT).
\citet{calixto-etal-2019-latent} propose a latent variable model for multi-modal MT, to learn an association between an image and its target language description.
\citet{long-etal-2021-generative,li_valhalla_2022} first synthesize an image conditioned on the source sentence, then use both the source sentence and the synthesized image to produce translation.
\citet{caglayan-etal-2020-simultaneous} obtain a lower latency in simultaneous MT by supplying visual context.
Differently, Vokenization \citep{tan-bansal-2020-vokenization} extend BERT \citep{devlin-etal-2019-bert} by implementing visual token prediction objective to learn a mapping between tokens and associated images. Most relevantly, \citet{lu_imagination-augmented_2022} improve text-only language understanding performance in low-resource settings by using generated imagination as visual supervision.

\section{Conclusion}
\label{sec:conclusion}
In this paper,
we described our two-stage method for the euphemism detection task.
We first collected literal descriptions for PETs, inserted these descriptions into the model input, and showed that such linguistic supervision greatly boosts performance. We then supplied extra visual supervision using a text-to-image model, where we denote this kind of supervision as visual imageries. We achieved a statistically significant performance increase by using visual imageries in addition to the term descriptions. Our qualitative analysis on visual imageries also suggests that a text-to-image model can be a functional tool to break down how models interpret figures of speech.

\noindent \textbf{Limitations.} Due to working with a small-scale dataset, we were able to manually collect descriptions for the PETs. Collecting these descriptions using an automatic retrieval system would be more sophisticated. We also did not perform a detailed analyses of the results, which could help shed light on the contribution of each model component.

\noindent \textbf{Acknowledgements}.
This work was supported in part by an AI Fellowship to I. Kesen provided by the KUIS AI Center, GEBIP 2018 Award of the Turkish Academy of Sciences to E. Erdem, and BAGEP 2021 Award of the Science Academy to A. Erdem. This publication is based upon work from COST Action \href{https://multi3generation.eu/}{Multi3Generation} CA18231, supported by \href{https://www.cost.eu/}{COST} (European Cooperation in Science and Technology).

\bibliography{anthology,custom}
\bibliographystyle{acl_natbib}

\appendix

\end{document}